\documentclass[manuscript,nonacm]{acmart}

\AtBeginDocument{%
  }

\acmJournal{CSUR}

\usepackage[normalem]{ulem}
\usepackage{multirow}
\usepackage{svg}
\usepackage[table,xcdraw]{xcolor}
\usepackage{enumitem}
\usepackage{color,soul}
\usepackage{booktabs}
\usepackage{caption}
\usepackage{pifont}

\newcommand{\cmark}{\ding{51}}%
\newcommand{\xmark}{\ding{55}}%

\begin{document}

\title{Learning from Observation: A Survey of Recent Advances}

\author{Returaj Burnwal}
\email{returaj.burnwal@gmail.com}
\affiliation{%
  \department{Department of Computer Science and Engineering}
  \institution{Indian Institute of Technology, Madras}
  \city{Chennai}
  \state{Tamil Nadu}
  \country{India}
}

\author{Hriday Mehta}
\email{hridaym.211me129@nitk.edu.in}
\affiliation{%
  \institution{National Institute of Technology, Karnataka}
  \city{Surathkal}
  \state{Karnataka}
  \country{India}
}

\author{Nirav Pravinbhai Bhatt}
\email{niravbhatt@dsai.iitm.ac.in}
\affiliation{%
  \department{Department of Data Science and AI}
  \institution{Indian Institute of Technology Madras}
  \city{Chennai}
  \state{Tamil Nadu}
  \country{India}
}
\affiliation{%
  \institution{Wadhwani School of Data Science \& AI}
  \city{Chennai}
  \state{Tamil Nadu}
  \country{India}
}

\author{Balaraman Ravindran}
\email{ravi@dsai.iitm.ac.in}
\affiliation{%
  \department{Department of Data Science and AI}
  \institution{Indian Institute of Technology Madras}
  \city{Chennai}
  \state{Tamil Nadu}
  \country{India}
}
\affiliation{%
  \institution{Wadhwani School of Data Science \& AI}
  \city{Chennai}
  \state{Tamil Nadu}
  \country{India}
}

\renewcommand{\shortauthors}{Burnwal et al.}

\begin{abstract}
Imitation Learning (IL) algorithms offer an efficient way to train an agent by mimicking an expert's behavior without requiring a reward function. IL algorithms often necessitate access to state and action information from expert demonstrations. Although expert actions can provide detailed guidance, requiring such action information may prove impractical for real-world applications where expert actions are difficult to obtain. To address this limitation, the concept of learning from observation (LfO) or state-only imitation learning (SOIL) has recently gained attention, wherein the imitator only has access to expert state visitation information. In this paper, we present a framework for LfO and use it to survey and classify existing LfO methods in terms of their trajectory construction, assumptions and algorithm's design choices. This survey also draws connections between several related fields like offline RL, model-based RL and hierarchical RL. Finally, we use our framework to identify open problems and suggest future research directions.
\end{abstract}




\maketitle

\section{Introduction}
\label{sec:introduction}
Imitation learning, often studied as part of reinforcement learning (RL), has emerged as a promising paradigm for training autonomous agents to perform tasks effectively by mimicking demonstrator  behavior \citep{il_survey}. In contrast to RL, which relies on environment reward, IL has the distinct advantage of learning solely from expert guidance, which alleviates the need to design a reward function. This feature proves particularly beneficial in complex environments where crafting a suitable reward function can be challenging. Standard methods developed in this framework often require access to demonstrations, in the form of state-action $(s_t, a_t)$ pairs. While expert actions can provide detailed guidance, requiring such action information may be restrictive in practical scenarios. For instance, a human expert may demonstrate the object-manipulation task to a robot by manually moving the robot's arm. The robot can record both its joint angles (state) and joint torques (action) induced by the human expert. Collecting such state-action pair demonstrations can be challenging or, in some cases, impossible; such as environments with limited access to physical demonstrators. 

Learning from demonstration (LfD) aims to replicate the expert's actions for a given state; this introduces a constraint that both the expert and the imitator must share the same dynamics model: specifically, they must have identical action spaces and the exact next-state transition probabilities for all feasible state-action pairs. This assumption brings severe limitations, as imagine that a robot with a low-speed limit imitates another robot that moves fast; then, using the LfD framework, a slow robot cannot learn to mimic the behavior of the fast-moving robot. The LfD framework does not support transfer learning across dynamically different agents \citep{sail}.

While IL research is inspired by the way humans learn from observations of another human, there exists a notable difference in the imitation process employed by humans. Unlike LfD, which often relies on demonstrator's action information to guide behavior, humans typically imitate without requiring explicit action information. A more general framework, known as learning from observation (LfO) or state-only imitation learning (SOIL), involves an expert communicating solely the state sequence information using raw sequences of true agent states, or partial observations like images or videos. In comparison to LfD, LfO aligns more closely with the natural way humans learn to imitate in the physical world, a phenomenon referred to as observational learning in psychology \citep{social_lt}. Moreover, the LfO framework can support transfer learning across dynamically different agents, as it does not rely on demonstrator's action information.

The recent works on LfO have demonstrated significant success in both simulated environments and real-world tasks, however the literature lacks a systematic review of this field. Though a novel taxonomy for LfO was proposed by \citet{LfO_survey}, we found that the recent LfO algorithms either do not neatly fit or cannot be categorised based on the established taxonomy (see section \ref{sec:related_works} for a detailed discussion of related work). Hence, this article presents a survey of the literature based on the fundamental questions on trajectory dataset construction and algorithm's design choices.\\

\noindent {\bf Overview}

\noindent The goal of this article is to present a comprehensive overview of learning from observation and to provide an over-arching framework to formalize this class of methods. We aim to define classification criteria based on the trajectory data set's construction and algorithms' design choices. We address the following fundamental questions to structure our classification:
\begin{itemize}
    \itemsep0em 
    \item Who qualifies as an expert? An expert can be a human, a distinct dynamical agent from the imitator, or the same dynamical agent as the imitator. The formulation of the algorithm varies depending on the identity of the expert with respect to the imitator. (Section \ref{sec:lfo:who_expert})
    \item How are expert trajectories collected? These trajectories can be collected in two primary ways: i) in a first-person viewpoint, where the state trajectories are collected from the agent-centric viewpoint. Collecting first-person demonstrations can be challenging or, in some instances, impossible. A more natural way would be to collect demonstrations from a ii) third-person viewpoint. (Section: \ref{sec:lfo:how_collect})
    \item What are the different trajectory datasets? In section \ref{sec:lfo:type_trajectory}, we survey different LfO algorithms based on the trajectory dataset they use to learn from expert demonstrations. A trajectory dataset can contain demonstrations from a single or multiple experts. Additionally, we provide a discussion of how the trajectory dataset varies depending on whether the algorithm operates online or offline. 
    \item How are algorithms formulated to facilitate learning from state-only demonstrations? In section \ref{sec:algo_learn}, we define and discuss our framework for classifying LfO approaches based on algorithm design choices.
\end{itemize}


In the next section, we will provide the necessary background for LfO algorithms. Section \ref{sec:traj_construction}, \ref{sec:algo_learn} forms the survey's core and will provide a comprehensive overview of the recent LfO methodologies. In Section \ref{sec:domain_adaptation}, we discuss how these algorithms are used to address learning across different viewpoints and non-identical agents. In Section \ref{sec:lfo_resources}, we discuss the datasets, benchmarks and algorithms used in LfO research. Following that, in Section \ref{sec:other_fields}, we will discuss the connections between LfO algorithms and other related fields in RL. Subsequently, in Section \ref{sec:open_problems}, we will use our framework to identify open problems and discuss future directions. Lastly, we will conclude the article by discussing related surveys in Section \ref{sec:related_works} and summarize this article in Section \ref{sec:summary}.
\section{Background}
\label{sec:backgound}
In this section, we will provide the necessary background for our work. We will begin by discussing the formulation of the RL and Markov Decision Process (MDP). Additionally, we will briefly discuss some approaches for LfD and introduce the notations used throughout our survey.

\subsection{Reinforcement Learning}

Reinforcement Learning focuses on how an autonomous agent should act within an environment to maximize a given reward signal \citep{rl_intro_book}. We formalize this interaction with the environment as Markov Decision Process (MDP). A MDP is defined by tuple $\mathcal{M} = (\mathcal{S}, \mathcal{A}, \mathcal{T}, r, \gamma, \rho_0)$, where $\mathcal{S}$ and $\mathcal{A}$ represents the state and action spaces, respectively. $\mathcal{T}(s' | s, a)$ defines the probability of transiting to state $s'$ after executing action $a$ in state $s$. Additionally, $r(s, a)$ denotes the immediate reward obtained from taking action $a$ in state $s$, while $\gamma$ stands for the discount factor. $\rho_0$ is the initial state distribution. 

Any policy, $\pi: \mathcal{S} \times \mathcal{A} \rightarrow [0, 1]$, induces a stationary state-action distribution $\mu^\pi(s,a)$, also called occupancy measure (See Table \ref{table:distribution}). \cite{mdpHandbook} showed that if $\mu$ satisfies the \textit{Bellman flow constraints}:
\begin{align}
    \label{eq:bellman_flow}
    &\sum_a \mu(s, a) = (1-\gamma) \rho_0(s) + \gamma \sum_{s',a'} \mathcal{T}(s | s', a') \mu(s', a')  \nonumber \\
    & \text{and } \mu(s,a) \geq 0 \;\;\; \forall\; s,a
\end{align}
Then, there exists a unique policy, $\pi_\mu (a | s) = \dfrac{\mu(s,a)}{\sum_a \mu(s, a)}$, whose occupancy measure is defined by $\mu$. Conversely, if a policy $\pi$ induces an occupancy measure of $\mu$, then $\pi = \pi_\mu$. Table \ref{table:distribution} summaries the other stationary distributions. The mentioned concept is principal to an important class of algorithms explored in Section \ref{sec:algo_learn:distributon_match}  which are used for solving LfO.

An agent follows a stochastic policy $\pi$ to interact with the MDP. The goal of the agent is to learn the optimal policy $\pi^*$  that maximizes the expected sum of discounted reward ($J_{RL}$).
\[
    \pi^* = arg\max_\pi J_{RL} := \mathop{\mathbb{E}}_{s_0, a_0, ..}\left[\sum_{t=0}^\infty \gamma^t r(s_t, a_t)\right] = \mathop{\mathbb{E}}_{(s, a) \sim \mu^\pi(s, a)} \left[r(s, a)\right]
\]

\noindent where, $\ s_0 \sim \rho_0,\ a_t \sim \pi(a_t | s_t),\ s_{t+1} \sim \mathcal{T}(s_{t+1} | s_t, a_t)$ and $\mu^\pi(s, a)$ is called the stationary state-action distribution under policy $\pi$.


\begin{table}[!tp]
\begin{center}
  \addtolength{\tabcolsep}{-2pt}
  \fontsize{9.0pt}{10.25pt}\selectfont
  \captionsetup{justification=centering}
  \caption{\small Definition of stationary distribution, \\where $\mu^\pi_t(s) = p(s_t = s | s_0 \sim \rho_0, a_i \sim \pi(a_i | s_i), s_{i+1} \sim \mathcal{T}(s_{i+1} | s_i, a_i)\ \forall\ i<t)$ }
  \label{table:distribution}
    \begin{tabular}{|c|c|c|c|c|c|}
    \hline
               & State & State-Action & Joint & State Transition & Inverse Dynamics \\ \hline
    Support    & $\mathcal{S}$  & $\mathcal{S}\times \mathcal{A}$  & $\mathcal{S}\times \mathcal{A} \times \mathcal{S}$ &  $\mathcal{S}\times \mathcal{S}$  & $\mathcal{S}\times \mathcal{A} \times \mathcal{S}$  \\ \hline
    Symbol     & $\mu^\pi(s)$ &  $\mu^\pi(s, a)$  &  $\mu^\pi(s, a, s')$  &  $\mu^\pi(s, s')$ &  $\mu^\pi(a| s, s')$ \\ \hline
    Definition &  $(1-\gamma) \sum_{t=0}^\infty \gamma^t \mu^\pi_t(s)$ &  $\mu^\pi(s)\pi(a|s)$  & $\mu^\pi(s, a) \mathcal{T}(s'|s, a)$ & $\int_\mathcal{A} \mu^\pi(s,a,s') da$ & $\dfrac{\mu^\pi(s,a,s')}{\mu^\pi(s,s')}$  \\ \hline
    \end{tabular}
\end{center}
\end{table}

Typically, the transition probability $\mathcal{T}$ is unknown in the environment. However, we assume it is possible to acquire samples $(s,a,s')$ by interacting with the environment. In situations where the state-action space is large, we employ a parameterized policy, $\pi(a|s; \theta)$, and seek to find the optimal policy within this representation. One such approach to find the optimal policy is known as the \textbf{Policy Gradient} (PG) method, wherein the policy parameters are directly optimized \citep{pg}. The policy gradient update equation is defined as:
\[
    \theta \leftarrow \theta + \alpha \nabla_\theta J_{PG}
\]
\[
    \nabla_\theta J_{PG} = \dfrac{1}{1-\gamma}\mathop{\mathbb{E}}_{(s,a) \sim \mu^\pi}\left[Q^\pi(s, a)\nabla \log(\pi(a | s; \theta
    ))\right]
\]

\noindent We define the action-value function $Q^\pi(s,a)$ as the expectation of the cumulative return under a policy $\pi$.
\[
    Q^\pi(s,a) = \mathop{\mathbb{E}}_{\pi, \mathcal{T}}\left[ \left.\sum_{t=0}^\infty \gamma^t r(s_t, a_t)  \right\vert s_0 = s, a_0 =a\right]
\]

\subsection{Imitation Learning}
In imitation learning, we rely not on environmental rewards,  $\mathcal{M} \symbol{92} r$, MDP excluding rewards, but on expert demonstrations to learn a behavior policy. However, multiple works
\citep{ridm,23_soil_iros_2021,mahalo_icml_2023,matt_rec1} use environmental rewards with demonstrations to learn a better behavior policy in a sample efficient manner. In this work, we do not survey these approaches and leave it for future works to discuss them in detail. To include the differences in dynamics, morphology or viewpoint of the expert and the imitator, we define expert MDP as $\mathcal{M}^E$ and imitator MDP as $\mathcal{M}^I\symbol{92} r$. The demonstrations can consists of state-action pairs $\tau^E \sim (s_0, a_0, .., s_t, a_t)$ or a state transition pairs $\tau^E_S \sim (s_0, s_1, .., s_t, s_{t+1})$. Additionally, we express the occupancy measure of the demonstrator $\mu^E$. 

\subsubsection{Learning from Demonstration}
In learning from demonstration, the agent requires expert demonstrations $\tau^E$, in the form of state-action pairs, and uses this demonstration to learn a behaviour akin to that of the expert. 

\noindent \textbf{Behaviour Cloning} (BC) uses supervised learning method to learn the imitator policy on $\tau^E$ \citep{bc}. This method, although simple, yields considerable efficacy and does not need additional interactions with the environment. Mathematically, BC based algorithm aims to minimize the KL divergence between the expert and the imitator policies.
\begin{align*}
    \min_\pi J_{BC} :&= D_{KL}(\pi^E(a|s)\ ||\ \pi(a | s))\\
                    &\propto \mathop{\mathbb{E}}_{(s,a) \sim \mu^E}[ - \log \pi(a | s)] \\ 
                    &\approx \mathop{\mathbb{E}}_{(s,a) \sim \tau^E}[ - \log \pi(a | s)]
\end{align*}
The BC-based policy is very fragile. When the BC policy interacts with the environment and if it makes an error in predicting expert action, it can lead the agent to explore a different state space not encountered in the expert demonstrations. As the imitator policy is trained through supervised learning, it may hallucinate actions for states that lack representation in the demonstrations. Consequently, the behavior of the imitator will diverge from that of the expert. This issue is also called the test-time shift problem in the literature \citep{bc_covariate}. One can address this challenge if BC policy is trained on ample expert demonstrations \citep{sasaki2018sample}. However, collecting a substantial amount of expert demonstrations can be costly.\\

\noindent \textbf{Inverse Reinforcement Learning} (IRL) is another class of algorithms that seek to recover the reward function responsible for generating the expert demonstrations $\tau_E$ \citep{irl_ng}. The recovered reward function is then used by a RL method to find an imitator policy. However, the main challenge of IRL stems from the fact that it is an ill-defined problem. There can be multiple optimal policies that can explain the expert demonstrations and many reward functions that can explain an optimal policy. To address the former ambiguity, the maximum entropy framework introduced by \citet{max_ent} selects a solution that makes minimal assumptions about the policy, making it as random as possible. \\

\noindent \textbf{Adversarial Imitation Learning} (AIL), tries to minimize the divergence of the stationary state-action distribution induced by the imitator and the expert policy. Consider the representation work of GAIL \cite{gail} where the policy $\pi$ is learned by minimizing the inverse KL divergence between the expert and the imitator's occupancy measure.  Mathematically, GAIL minimizes the following objective.
\begin{align}
    \label{eq:gail_1}
    &\min_\pi D_{KL}(\mu^\pi(s,a) \;||\; \mu^E(s,a)) = \mathbb{E}_{(s,a)\sim \mu^\pi}\left[ \log\left( \dfrac{\mu^\pi(s,a)}{\mu^E(s,a)} \right) \right] \nonumber \\
    \text{equivalently, } &\max_\pi -D_{KL}(\mu^\pi(s,a) \;||\; \mu^E(s,a)) = \mathbb{E}_{(s,a)\sim \mu^\pi}\left[ \log\left( \dfrac{\mu^E(s,a)}{\mu^\pi(s,a)} \right) \right]
\end{align}
The above maximization equation, Eq \ref{eq:gail_1}, is convenient, as it can be reformulated as an RL optimization problem, where the reward is given by the density ratio.
\begin{align}
    \label{eq:gail_2}
    \max_\pi \mathbb{E}_{(s,a)\sim \mu^\pi}\left[ \log\left( \dfrac{\mu^E(s,a)}{\mu^\pi(s,a)} \right) \right] = (1-\gamma) \mathbb{E}_{\pi,\mathcal{T}} \left[ \sum_{t=0}^\infty \gamma^t \log\left( \dfrac{\mu^E(s,a)}{\mu^\pi(s,a)} \right) \right]
\end{align}
In simpler terms, having access to the estimates of the density ratios for the two policies can transform the divergence minimization problem into a standard RL problem, with the reward function defined as $r(s,a) = \log\left( \dfrac{\mu^E(s,a)}{\mu^\pi(s,a)} \right)$. As we lack direct access to the density estimates of $\mu^E$ and $\mu^\pi$, we estimate the density ratio by training a classifier (discriminator) model using the samples from $\mu^E$ and $\mu^\pi$, formulated as follows:
\begin{align}
    \label{eq:gail_disc}
    \max_{c:\mathcal{S}\times\mathcal{A}\rightarrow (0,1)} \mathbb{E}_{(s,a)\sim \mu^E}[\log(c(s,a))] + \mathbb{E}_{(s,a)\sim \mu^\pi}[\log(1-c(s,a))]
\end{align}
In the above objective, the optimal value of $c^*(s,a) = \dfrac{\mu^E(s,a)}{\mu^E(s,a) + \mu^\pi(s,a)}$. We can, therefore, estimate the log-density ratio using the optimal $c^*$ as:
\begin{align}
    \label{eq:log_density}
     r(s,a) = \log\left( \dfrac{\mu^E(s,a)}{\mu^\pi(s,a)} \right) = \log(c^*(s,a)) - \log(1-c^*(s,a))
\end{align}
We can then use the above reward signal to update the policy using some RL algorithm. However, in practice, the discriminator is not fully optimized; instead, updates are made to both the discriminator and the policy network alternately.\\ 

\noindent \textbf{Distribution Correction Estimation algorithms} (DICE) \citep{valuedice, optidice, demodice, relaxdice} focus on optimizing the difference between the stationary distribution within the distribution space rather than directly optimizing the policy. Subsequently, the estimated distribution is used to train the imitator policy.


\subsubsection{Learning from Observation}
LfO poses a more challenging problem, as expert demonstrations are available in the form of state transition sequences $\tau^E_S \sim (s_0, s_1, ...)$. In the following sections we will discuss the underlying assumptions, similarities and differences among the LfO approaches in detail. 


\subsection{Forward and Inverse Dynamics Model}
A forward dynamics model ($\mathcal{S} \times \mathcal{A} \rightarrow \mathcal{S}$) predicts the next state given the current state and action. In RL literature, it is either used as a proxy for the environment or for look-ahead planning. On the other hand, an inverse dynamics model ($\mathcal{S} \times \mathcal{S} \rightarrow \mathcal{A}$) predicts the one-step action needed to transition from the current to the next state.
\begin{table}[!tp]
  \centering\small
  \addtolength{\tabcolsep}{-4pt}
  \captionsetup{justification=centering}
  \caption{\label{table:traj_construction}Overview of the taxonomy for trajectory construction.}
\begin{tabular}{@{}lll@{}}
\toprule
\rowcolor[HTML]{C0C0C0} 
\textbf{Dimension}                                & \textbf{Consideration}                                                                                                                           & \textbf{Key Ideas}                                                                                                                                                                 \\ \midrule
                                                  & Same as imitator                                                                                                                                 & \begin{tabular}[c]{@{}l@{}}\(\mathcal{M}^E = \mathcal{M}^I\) \\ Shares the same MDP\end{tabular}                                                                                   \\ \cmidrule(l){2-3} 
                                                  & Dynamically different                                                                                                                            & \begin{tabular}[c]{@{}l@{}}\(\mathcal{T}^E \neq \mathcal{T}^I\) \\ Shares the same state, action space and \\ initial state distribution\end{tabular}                              \\ \cmidrule(l){2-3} 
\multirow{-3}{*}{1. Expert type}                  & Morphologically different                                                                                                                        & \begin{tabular}[c]{@{}l@{}}\(\mathcal{S}^E \neq \mathcal{S}^I\), \(\mathcal{A}^E\neq \mathcal{A}^I\),\\ \(\mathcal{T}^E \neq \mathcal{T}^I\), \(\rho_0^E = \rho_0^I\)\end{tabular} \\ \midrule
                                                  & First Person Viewpoint                                                                                                                           & \begin{tabular}[c]{@{}l@{}}\(\mathcal{S}^E = \mathcal{S}^I\)\\ May not share the same action space or\\ transition dynamics model\end{tabular}                                     \\ \cmidrule(l){2-3} 
\multirow{-2}{*}{2. Expert trajectory collection\;\;\;} & Third Person Viewpoint                                                                                                                           & \begin{tabular}[c]{@{}l@{}}\(\mathcal{S}^E \neq \mathcal{S}^I\)\\ May not share the same action space or \\ transition dynamics model\end{tabular}                                 \\ \midrule
                                                  & Demonstrations from single expert                                                                                                                &                                                                                                                                                                                    \\ \cmidrule(l){2-3} 
                                                  & \begin{tabular}[c]{@{}l@{}}Demonstrations from multiple experts \\ with varying performances\end{tabular}                                        &                                                                                                                                                                                    \\ \cmidrule(l){2-3} 
                                                  & \begin{tabular}[c]{@{}l@{}}Demonstrations from multiple experts\\ with varying skills\\ (distinct patterns of behaviours)\end{tabular}           &                                                                                                                                                                                    \\ \cmidrule(l){2-3} 
                                                  & \begin{tabular}[c]{@{}l@{}}Demonstrations from expert with \\ additional proxy state‑trajectories \\ from imitator and expert agent\end{tabular} & \begin{tabular}[c]{@{}l@{}}Used in third person IL. \\ Additional trajectories are used to \\ align the state space of the expert\\ and the imitator agent.\end{tabular}           \\ \cmidrule(l){2-3} 
\multirow{-5}{*}{3. Type of trajectory dataset}   & \begin{tabular}[c]{@{}l@{}}Demonstrations from expert with \\ additional state‑action trajectories \\ from imitator agent\end{tabular}           & \begin{tabular}[c]{@{}l@{}}Additional state‑action trajectories \\ from imitator agent is required for\\ offline learning.\end{tabular}                                            \\ \bottomrule
\end{tabular}
\end{table}

\section{Trajectory dataset construction for LfO}
\label{sec:traj_construction}
To properly discuss LfO algorithms, we first need to understand how our given trajectory datasets are constructed. In this section, we will discuss the construction of the trajectory dataset. We will specify a taxonomy that disentangles three central questions of trajectory construction. The three main questions we answer are:

\begin{enumerate}
    \item Who qualifies as an expert? (Section \ref{sec:lfo:who_expert})
    \item How are expert trajectories collected? (Section \ref{sec:lfo:how_collect}) 
    \item What are the different trajectory datasets? (Section \ref{sec:lfo:type_trajectory})
\end{enumerate}

The first question focuses on the expert's identity, which is used to collect demonstrations for imitation, while the last two questions cover the trajectory dataset construction. There are several important considerations in each of the above questions, leading to the overall taxonomy summarized in Table \ref{table:traj_construction}. In the following subsections, we elaborate on each of these considerations. 

\subsection{Who qualifies as an expert?}
\label{sec:lfo:who_expert}
Standard IL requires demonstrations in the form of state-action pairs, which limits the imitator agent to sharing the same morphology, dynamics, and viewpoint as the expert. However, in LfO, action information is not necessary in the demonstrations. This allows for demonstrations from various types of experts, as long as they share either partial state space \citep{1_third_person_il_iclr_2017, 15_opolo_neurips_2020, 23_soil_iros_2021, 18_cross_domain_il_icml_2021, 24_smodice_icml_2022} or behavior level information \citep{10_decoupled_hierarchical_il_neurips_2019, 18_cross_domain_il_icml_2021, 20_disentanGAIL_iclr_2021, 39_domain_adaptive_IL_neurips_2023} with the imitator agent. 
We categorize the expert as follows:
\begin{enumerate}[label=\roman*]
    \item \textit{Same as imitator agent}: An obvious expert is an agent that shares the same dynamics as the imitator agent  \citep{1_third_person_il_iclr_2017, 2_imitation_from_raw_video_icra_2018, 4_bco_ijcai_2018, 5_trex_icml_2019, 9_gaifo_icml_workshop_2019, 15_opolo_neurips_2020,23_soil_iros_2021, 27_depo_icml_2022, 28_lobsdice_neurips_2022}. In this scenario, the imitator agent can replicate the exact state-only demonstrations provided by the expert. Consequently, algorithms \citep{9_gaifo_icml_workshop_2019,15_opolo_neurips_2020, 27_depo_icml_2022} in this category typically aim to align the imitator's state-transition with that of an expert, thereby ensuring a close match between their state-trajectory outcomes. 

    \item \label{who_expert:dynamics} \textit{Dynamically different from imitator agent}: Since LfO does not require action information in its demonstrations, we can collect state-only demonstrations from dynamically distinct expert agents that share the same state space with the imitator agent. In the real world, even if the expert and the imitator robots are of the same make, assuming that the system dynamics are the same is not practical. Changes to a robot's dynamics can be due to internal changes, such as mechanical faults \citep{verma2004real}, dropping battery charge-level \citep{hutter2017anymal}, and external changes, such as changes in the operating environment, e.g., surface friction \citep{hao2021dynamic}, or the robot's task, e.g., increased load \citep{hutter2017anymal}. In such cases, one cannot assume that the imitator agent can follow the exact state trajectory sequence demonstrations. Therefore, algorithms \citep{12_sail_iclr_2020, 14_i2l_iclr_2020, 24_smodice_icml_2022} in this category train the imitator agent by matching the stationary state distribution ($\mu^\pi(s)$) of the expert agent. Intuitively, the imitator agent can replicate expert behavior if both agents have similar state visitation probability. It is important to note that the matching state visitation does not force the imitator agent to share the same next-state transition distribution with the expert agent. Matching state visitation has had empirical success but has a severe limitation. Consider a scenario where the expert moves in a clockwise direction in a circle, and the imitator learns a policy to move in a counterclockwise direction. In this case, the imitator and expert share the exact state-visitation probabilities; however, the imitator agent does not behave similarly to the expert. An alternate approach, as suggested in \citet{50_fmdp_2021,51_imperfect_demo_icra_2021,31_ailo_iclr_2022,33_ood_il_corl_2023}, is to identify all feasible state transitions from demonstrations and then minimize the state-transition distribution.

    \item \textit{Morphologically different from imitator agent}: This category discusses the scenario where the morphologically different expert provides demonstrations. These demonstrations hold enough information to transfer the expert’s behavior to an imitator agent. For instance, consider a navigation task where a wheeled robot is learning to imitate from the demonstrations of a legged robot or humans. Given that the demonstrations come from distinct agents, the challenge lies in effectively transferring behavior information to the imitator agent. \citet{18_cross_domain_il_icml_2021, 10_decoupled_hierarchical_il_neurips_2019, 30_udil_neurips_2022} approach this imitation problem by learning a domain mapping function. This function maps the state representation of the expert (source domain) to the state representation of the imitator (target domain). The imitator agent then learns to imitate the mapped state representation. \citet{42_tcn_icra_2018, 20_disentanGAIL_iclr_2021, 39_domain_adaptive_IL_neurips_2023} proposes an alternate approach, where it learns a domain-invariant feature model, which maps both expert and imitator state into the same feature space. The expert feature representation is then used to learn the imitator policy.

\end{enumerate}

\subsection{How are expert trajectories collected?}
\label{sec:lfo:how_collect}

In LfO, we collect state-only information in our demonstrations. In this section, we will focus on different ways of collecting state information. This state information can be collected from two perspectives: first-person and third-person viewpoint. 
\begin{enumerate}[label=\roman*]
    \item \textit{First-person viewpoint}: The first-person viewpoint is the most common way for collecting state-only demonstrations from the expert. In this perspective, expert demonstrations are collected so that the expert and the imitator observe the state space from the same viewpoint. The imitator agent’s goal is to learn to traverse the same sequence of states as demonstrated by the expert. \citep{4_bco_ijcai_2018, 27_depo_icml_2022} present an algorithm that learns an inverse dynamics model for the imitator agent and uses this model to estimate actions for given expert state transitions. On the other hand, \citep{9_gaifo_icml_workshop_2019, 12_sail_iclr_2020, 28_lobsdice_neurips_2022} propose an alternate approach where the imitator agent learns to imitate by matching the stationary state or state-transition distribution. Though first-person demonstrations can be easier to learn for an imitator agent, it may not be practical or sometimes impossible to obtain first-person demonstrations in the real world.
    
    \item \textit{Third-person viewpoint}: The third-person viewpoint is a more practical way of collecting demonstrations in the real world. This perspective mirrors how humans learn to imitate by observing others perform tasks from a third-person viewpoint. In this context, the state information from the expert may partially align with the state space of the imitator agent. However, the expert's demonstration offers enough insight for the imitator to learn and replicate the expert's behavior. \citet{1_third_person_il_iclr_2017, 42_tcn_icra_2018, 45_tdc_cmc_neurips_2018, 20_disentanGAIL_iclr_2021, 39_domain_adaptive_IL_neurips_2023} proposes to learn features independent of the viewpoint, which can then be used to mimic the expert's behavior. An alternative method, as depicted in several works \cite{2_imitation_from_raw_video_icra_2018, 10_decoupled_hierarchical_il_neurips_2019, 18_cross_domain_il_icml_2021, 30_udil_neurips_2022}, is to learn a mapping function that can translate the state representation from the expert's viewpoint to the imitator's, followed by applying first-person imitation techniques.
\end{enumerate}

We present more detailed discussion on these algorithms and their key differences in section \ref{sec:algo_learn}. In existing literature, state demonstrations from a first-person viewpoint are typically collected as images or system-level information (such as sensor data). However, most third-person viewpoint studies collect demonstrations in the form of images.

\subsection{What are the different trajectory datasets?}
\label{sec:lfo:type_trajectory}

The above two sections discussed dataset construction based on the expert identity and how the state demonstrations are collected. In this section, we explore the different types of demonstration datasets: 

\begin{enumerate}[label=\roman*]
    \item \textit{Demonstrations from single expert}: A standard method for constructing dataset is by collecting expert demonstrations from a single expert \citep{1_third_person_il_iclr_2017, 4_bco_ijcai_2018, 11_ilpo_icml_2019, 15_opolo_neurips_2020, 12_sail_iclr_2020, 23_soil_iros_2021, 27_depo_icml_2022}. In this process, we collect a few or multiple state-only demonstrations using one expert. As these demonstrations originate from a single expert, they can be regarded as samples from a particular expert state-trajectory distribution. In other words, these demonstrations exhibit similar behavior, and the goal of the imitator agent is to align with the expert state-trajectory distribution.

    \item \textit{Demonstrations from experts with varying performance}: A demonstration dataset can also be constructed using experts whose performances range between high and low, unlike learning from single-expert demonstrations, which require optimal demonstrations to behave optimally in the environment. Sub-optimal demonstrations are more straightforward to collect. The imitator agent’s objective is then to match or surpass the performance of the best demonstration. However, it still requires all the experts to share the same morphology or MDP. \cite{5_trex_icml_2019, 53_ranking_il_tmlr_2023} accomplished this by using ranked demonstrations from high to low-quality trajectories and then learning a reward model that explains the order of these demonstrations. Following this, a standard RL algorithm is trained using this reward model to imitate or occasionally exceed the performance of the best expert demonstration. 
    
    \item \textit{Demonstrations from multiple experts with varying skills}: Collecting demonstrations from a morphologically identical agent can be an expensive task \citep{zhu2022self}. We can address this challenge by utilizing existing demonstrations from other agents with different skills \citep{12_sail_iclr_2020, 14_i2l_iclr_2020, 20_disentanGAIL_iclr_2021, 50_fmdp_2021, 51_imperfect_demo_icra_2021, 31_ailo_iclr_2022, 30_udil_neurips_2022, 33_ood_il_corl_2023}. The goal of the imitator agent is to learn to utilize different skills (such as crawling, walking and trotting) to mimic the experts' behavior. Learning from demonstrations of morphologically different agents is closer to how humans learn to imitate in real world. 

    \item \textit{Demonstrations from expert and proxy task state-trajectories from both expert and imitator agent}: Works \citep{ 18_cross_domain_il_icml_2021, 20_disentanGAIL_iclr_2021} requiring additional proxy state-trajectories from both the expert and the imitator agents learn the imitator policy from third-person viewpoint demonstrations. They use these proxy trajectories to learn to map the behaviors between the agents. Using proxy trajectories is one way to learn from demonstrations from different viewpoints. We discuss these approaches in more detail in section \ref{sec:third_person_il}.
    
    \item \textit{Demonstrations from expert and arbitrary state-action trajectories from imitator agent}: In this scenario, the dataset comprises the expert's state-only demonstrations and additional arbitrary state-action trajectories from the imitator agent. LfO algorithms \citep{28_lobsdice_neurips_2022, 24_smodice_icml_2022} that utilize this type of dataset typically learn to mimic expert behavior in an offline manner. Including additional state-action trajectories from the imitator agent serves two distinct purposes. As online interactions with the imitator agent are not allowed, the imitator policy can use this additional trajectory to learn the effect of the action on the imitator agent. It also helps mitigate offline IL's test-time distribution shift problem, as outlined by \cite{bc_ross}.

\end{enumerate}

\section{Algorithmic design choices for LfO}
\label{sec:algo_learn}
In the previous section, we discussed the choices related to expert trajectory construction. In this section, our primary focus is understanding the algorithmic design for LfO. We begin by classifying various LfO approaches, discussing their central ideas, and briefly discussing the approaches within each category. The question we seek to address is:

\begin{center}
    \textit{How are algorithms formulated to facilitate learning from state-only demonstrations?}
\end{center}

To answer the above question, we classify LfO approaches into four main categories and discuss their central idea:
\begin{enumerate}
    \item \textit{Supervised Approach}: In this category, the imitator policy undergoes training by directly optimizing a supervised loss function. (Section \ref{sec:algo_learn:supervised_approach})
    \item \textit{Goal from Expert Demonstrations}: In this category, the imitator policy is subdivided into two distinct sub-policies: meta-policy and low-level policy. The meta-policy selects a goal state from the expert demonstration, while the low-level policy tries to reach the defined goal state. (Section \ref{sec:algo_learn:expert_as_goal})
    \item \textit{Reward Engineering}: The reward function is derived from the expert demonstrations in this approach. Subsequently, the imitator policy is learned by optimizing this derived reward function. (Section \ref{sec:algo_learn:reward_engineering})
    \item \textit{Distribution matching}: Within this category, the imitator policy undergoes training via minimizing the divergence between the expert and imitator agents' stationary state or state-transition distributions. (Section \ref{sec:algo_learn:distributon_match})
\end{enumerate}
In the following sections, we will detail each of the categories.
\subsection{Supervised Approach}
\label{sec:algo_learn:supervised_approach}
Works in this category learn the imitator policy by minimizing the supervised objective function. \citet{4_bco_ijcai_2018} presents an algorithm that utilizes exploratory policy to interact with the environment and the collect trajectories that are then employed to train an inverse dynamics model. This model then estimates the actions for the given expert state and next-state pairs. Subsequently, a behavior-cloned policy is trained based on the expert state and the predicted action. The algorithm shares similarities with the BC-based algorithm, and as a result, it also encounters the same test-time shift problem \citep{bc_covariate}. Additionally, learning an inverse dynamics model requires a large number of interactions with the environment using an exploratory policy, which can be risky when the cost of interaction with the environment is high. To address the latter challenge, \citet{11_ilpo_icml_2019} proposes an alternate method. First, the policy is learned offline in a latent action space that describes the expert state transitions. Subsequently, an action mapping network aligns the latent policy to the actual action labels. The action mapping network can be learned through a small number of interactions with the environment. The latent policy network is learned by minimizing the error between the actual next state and the expected next state given by the latent policy. Nevertheless, the main challenge with this approach lies in its assumption that the action space can be represented using a discrete set of latent actions. This assumption may not apply to complex environments with large or continuous action spaces. \citet{29_ild_cvpr_2022} proposes to use a differentiable physics simulator for policy learning. It unrolls the dynamics by sampling action from the imitator policy. It then minimizes the distance between the expert and imitator trajectories by back-propagating the gradient into the imitator policy via a differentiable environment. The loss function contains deviation loss that measures the distance between all the imitator states and the closest expert states from the demonstration. This loss ensures that the imitator agent produces states closer to the expert state demonstration. However, the deviation loss may cause state collapse, as the imitator agent may only generate states closer to some expert state demonstration. In order to ensure the coverage of all the expert states, the authors introduce another term in the loss function, which measures the distance between all the expert states and the closest imitator state from the imitator trajectory. This loss function forms a single function that is easier to optimize.

One can efficiently use algorithms in this category to imitate similar agents in a simpler environment. These approaches may fail to recover expert behavior for more complex scenarios where either fewer expert demonstrations are available or environment interactions are limited.

\subsection{Goals from Expert Demonstration}
\label{sec:algo_learn:expert_as_goal}
\begin{figure}
    \centering
    \includegraphics[width=0.9\textwidth]{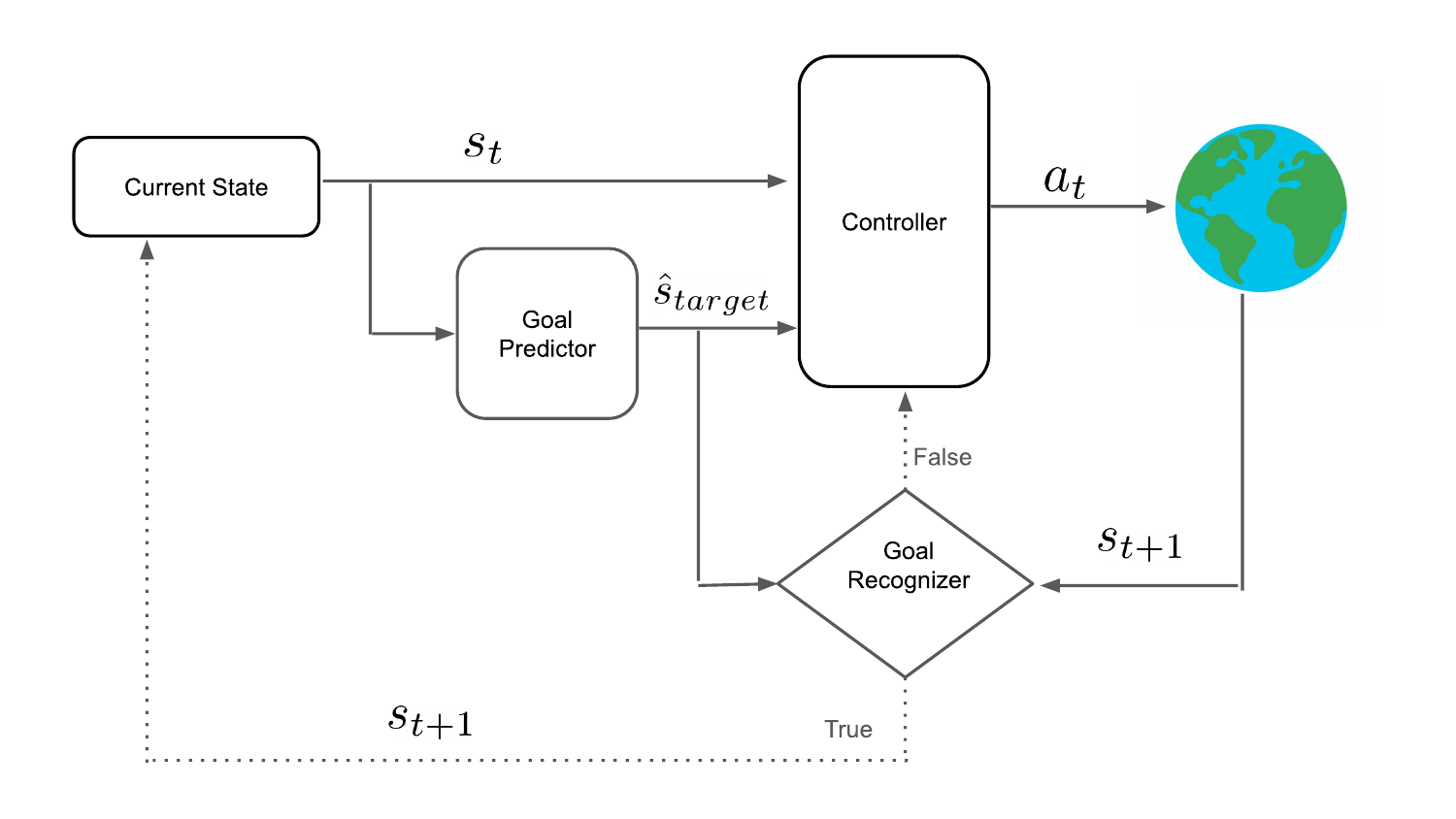}
    \Description{Diagram showing the process of selecting a goal from expert demonstration, with modules for goal prediction, low-level control, and goal recognition.}
    \caption{Goal from expert demonstration. The goal predictor estimates the goal state from the expert demonstrations. The low-level controller module predicts a sequence of actions needed to reach the this goal, following which, the next goal state is obtained. A goal recognizer evaluates the proximity of the current state and the goal, providing signals depending on if the current state is close to goal state.}
    \label{fig:goal_from_expert}
\end{figure}

The key idea of the approaches in this category is that an imitator agent should not only understand the demonstrator's intent (where to reach) but also execute the intended task (how to reach). This insight naturally leads to subdividing the imitator policy into two sub policy. The \textit{meta policy} is a target prediction module that predicts the goal state (where to reach) from the expert demonstration, and the \textit{low-level controller policy} outputs sequence of actions (how to reach) to reach the goal state, as shown in Figure \ref{fig:goal_from_expert}. The inferring of the goal state can be done in two ways:

\subsubsection{Selecting Goal from Expert Demonstration}
The approaches in this section define meta policy such that it selects the goal state directly from the expert demonstration, i.e., the chosen state is in the expert demonstration. The work by \citet{3_gsp_iclr_2018} suggests using a goal-conditioned controller that executes multiple actions until the agent reaches the desired goal state. Additionally, it trains a goal recognizer model to predict whether the goal state has been reached. The meta policy then selects the subsequent expert state in the demonstration as the next goal state for the imitator agent to reach. Therefore, the imitator agent learns to mimic the expert's behavior by visiting all the demonstrated states in the specified order. However, this approach inherently assumes that the imitator agent can visit all the demonstrated states, which may not hold in scenarios where the demonstration comes from a different agent. \citet{7_silo_corl_2019} proposes an approach that allows the imitator agent to skip the infeasible parts of the demonstration. The meta policy selects the goal state from the future states in the demonstration. The meta policy and the low-level controller both receive a positive reward signal if the low-level controller successfully reaches the goal state. They are then optimized using the standard RL technique.


\subsubsection{Estimating Goal from Expert Demonstration} 
The meta policy here predicts the goal state from the expert demonstration, which may not exactly match the states in the expert demonstration. \citet{10_decoupled_hierarchical_il_neurips_2019} proposes to decouple the goal estimate module from the controller. The low-level controller policy uses the imitator’s current and goal states to predict a sequence of actions. Since the controller policy is task-independent, it can be trained using any transitions. Meanwhile, the meta policy is trained to predict the goal state given the current imitator state and the demonstration. \citet{27_depo_icml_2022} proposes a similar approach to decouple the goal generator from the controller. They employ an inverse dynamics model as the controller,  which predicts actions based on the current imitator and goal states. The meta policy is trained using demonstrations to predict the goal state given the current imitator state. The meta policy undergoes additional updates through adversarial-style training. Decoupling these modules also facilitates knowledge transfer. For instance, when the imitator task or agent changes, one can retrain the goal generator module or the controller, respectively, and reuse the other module. The above works rely on robot demonstrations, the collection of which is resource-intensive and are required in large numbers for long-horizon, multimodal tasks. In contrast, human demonstrations are much cheaper and easier to collect than robot demonstrations. \citet{wang2023mimicplay} proposes to use human demonstrations to learn the meta policy.

Algorithms in this category decompose the imitation task into deciding \textit{where} to reach and \textit{how} to reach there. This design can aid the agent in imitating in complex environments. However, it is also prone to compounding errors that may arise from inaccuracies in estimating goal states.



\subsection{Reward Engineering}
\label{sec:algo_learn:reward_engineering}
\begin{figure}
    \centering
    \includegraphics[width=0.9\textwidth]{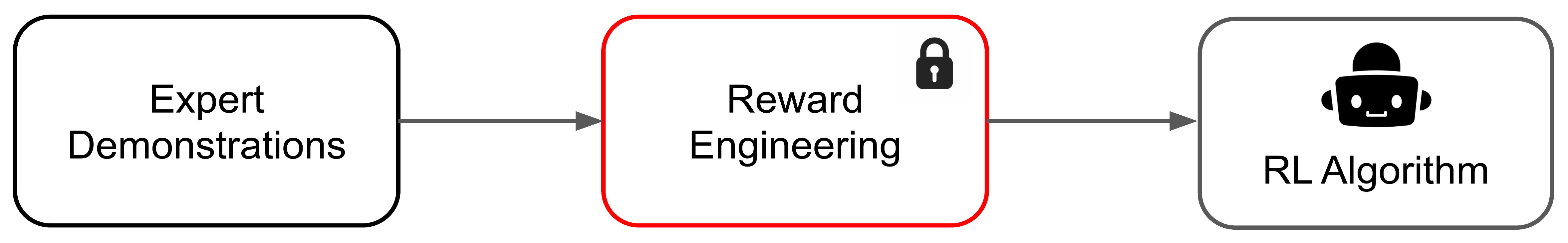}
    \Description{The diagram shows that the reward function is a fixed function of the expert demonstrations. Once the reward function is defined, a standard RL algorithm is used to learn the imitator policy.}
    \caption{Reward Engineering. A reward is defined as fixed function of expert demonstrations that takes input imitator agent state or state transition to output how close it is to the expert demonstrations. A standard RL technique is then used to learn the imitator policy based on this reward function. Works in this category differ in the choice of the metric they use to evaluate similarity to the expert demonstrations, which  influences the performance of the imitator policy.}
    \label{fig:rewards}
\end{figure}


In this category, the custom reward function is used to guide the learning of imitation policies using standard RL techniques, concisely depicted in Figure \ref{fig:rewards}. The reward function may not necessarily be the same as the one used to obtain expert demonstrations; instead, it is an estimate inferred from the demonstration dataset. \citet{2_imitation_from_raw_video_icra_2018} maps the expert and imitator state information into a common feature space where similar states have the same feature representation. The reward function then corresponds to minimizing the squared Euclidean distance between the encoding of the current imitator observation and the average of all the encoded state demonstration features. \citet{45_tdc_cmc_neurips_2018} also takes a similar approach to learn common feature representation. However, the reward function is defined as a dot product between the imitator's feature and the expert's checkpoint feature to measure the closeness between the observations. \citet{42_tcn_icra_2018}, on the other hand, defines the reward function as a squared Euclidean distance with Huber-style loss. The squared Euclidean distance gives us stronger gradients when the representations are further apart, whereas Huber-style loss prevails when the representation vectors are close. Alternatively, \citet{26_bootifol_neurips_workshop_2022} proposes to learn a sequence encoding function, which encodes a sequence of state observations of an agent trajectory. The reward function is the distance between the encoded expert and the imitator agents' trajectories until the current timestep. \citet{44_osl_icra_2019} proposes to train a model using a shuffle-and-learn style loss \citep{shuffle_and_learn_eccv_2016}. This loss function enables the model to learn the order of observations in the demonstration data. Subsequently, the reward function is defined to predict the task completion rate using the learned model. Another approach by \citet{19_form_icml_2021} learns next-state prediction models conditioned on previous states for both the expert and the imitator agent by sampling state transitions from demonstrations and imitator policy, respectively. The reward function is then defined as the log density ratio between the expert and the imitator next-state prediction models. However, learning such a state conditioned generative model for the expert requires prior offline training, which results to a less efficient and more time consuming approach. \citet{5_trex_icml_2019} learns the reward model from the performance-ranked expert demonstrations. Unlike previous approaches, which seek a reward function that explains the expert demonstrations, this approach learns a reward model to explain the ranking over the demonstrations. The reward model is trained using a binary classification loss, ensuring the high-performing trajectory yields greater returns than lower-performing ones. However, using such reward functions often tend to poor performances when considering tasks that operate in an inherent sequential manner. With rewards primarily associated with the later stages of the task, essential early skills are often not learnt which lead to inaccurately learned policies. For example, consider a task where a pen needs to be put in a box. The preliminary behaviour would be to pick up the pen, and place it in a box. However, due to rewards being focussed on the completion of the task in the latter stages, the agent can simply imitate going near the box without the pen, gaining the local reward signal. \cite{liu2023imitation} propose to use a dynamic discount factor to incentivize the agent to learn early (by keeping a low discount factor) as well as latter behaviors (by increasing the discount factor). This factor is quantified by formulating a progress recognizer to deduce the timesteps till which the agent has successfully imitated the expert. The reward function here is formalized as an optimal transport problem \cite{optimaltransport}.

Works under this category are easy to formalize. However, the performance of the algorithm relies on the choice of the reward engineering that is used to drive the learning of the imitation policy.

\subsection{Distribution Matching}
\label{sec:algo_learn:distributon_match}
In this category, the imitator policy is learned by aligning the stationary distribution induced by the expert and the imitator policy. The primary objective here is to minimize the divergence between the state transition ($\mu(s_t,s_{t+1})$) or the state ($\mu(s_t)$) occupancy measures. We make different assumptions about the imitator agent based on which occupancy measure we choose to minimize the divergence. For instance,  minimizing the state-transition occupancy measure assumes that the imitator agent can realize the same expert transition ($s^E_t, s^E_{t+1}$). However, if the expert and the imitator agent have distinct dynamics, there is no guarantee that the imitator agent can achieve the same expert transition. In such cases, one must identify all feasible transitions and align the state-transition distribution measure. Another way could be to use different objectives, such as minimizing the divergence of the state occupancy measure. However, as discussed in Section \ref{sec:lfo:who_expert}, matching state occupancy measures, in general, will not assure similar expert behavior. We can align occupancy measures using two different ways: Adversarial and DICE-based learning way.


\subsubsection{Adversarial Imitation Learning}
Works in this category extend the influential work of GAIL \citep{gail}. Primarily, it adopts a Generative Adversarial Networks (GAN, \cite{gan}) style strategy to perform distribution matching. A discriminator network is trained to distinguish between the state transition or the state generated from the imitator policy and the reference expert trajectory data distribution. The output of the discriminator serves as the feedback signal that encourages the imitator agent to generate similar behavior to the demonstration. We will now discuss the works based on matching the state transition and the state occupancy measures.

\begin{enumerate}[label=\roman*]
    \item \textit{Matching state transition occupancy measure}: Here, the objective of the imitator policy is to minimize the discrepancy between the state transition stationary distribution induced by the expert and the imitator policy. Mathematically, imitator policy minimizes the divergence (such as inverse KL divergence) between the state transition occupancy measures.
    \begin{align}
        \label{eq:state_transition_1}
        \min_\pi J_{s,s'} :&= D(\mu^\pi(s, s')\ ||\ \mu^E(s, s'))
    \end{align}
    GAIfO, introduced by \cite{9_gaifo_icml_workshop_2019}, follows a similar procedure of GAIL \citep{gail}. However, the crucial difference is that the discriminator is trained to classify state transition ($s_t, s_{t+1}$) rather than state action ($s_t, a_t$) pair between the imitator policy and expert data distribution. \citet{6_iddm_neurips_2019} identifies that the inverse dynamics disagreement ($D(\mu^\pi(a | s_t,s_{t+1}) || \mu^E(a|s_t,s_{t+1}))$) accounts for the gap between the GAIL and the GAIfO objective. It then optimizes the upper bound of the inverse dynamics disagreement with the GAIfO objective in a model-free way. The work of \citet{25_robust_gaifo_aamas_2022} focuses on learning a robust GAIfO policy in the simulator environment where the transition dynamics of the real environment and the simulator are slightly different. Additionally, the study assumes that the real and the simulator environments belong to the same fixed class of transition dynamics. The proposed algorithm learns a robust policy over this fixed class of transition dynamics while acting only in the simulator environment. The algorithms discussed are designed to learn an imitator policy tailored to the given expert state sequences. If there are changes in the expert's state trajectory, the imitator agent must relearn the policy from the beginning, as it cannot transfer knowledge from the previous task. The approach described in \citet{27_depo_icml_2022} separates the imitator policy into two main components: a high-level state planner responsible for generating the target state based on the current state and an inverse dynamics model that predicts the action based on the current state and the target state. The inverse dynamics model can be trained through any interactions with the environment and can be used to transfer knowledge. On the other hand, the state planner is trained to predict the expert's next state in a supervised manner, and its performance is further improved through GAIfO-like training. One of the primary limitations of GAIfO-style training is its reliance on on-policy samples. This restriction prevents effective utilization of previously collected transitions from the imitator policy during training. In an attempt to address this challenge, \citet{15_opolo_neurips_2020} presents an off-policy training strategy. The algorithm optimizes the upper bound of the equation \ref{eq:state_transition_1} objective by involving previously collected transitions. In this method, the discriminator classifies whether the state transitions originate from the expert or the past imitator policy samples. The feedback from the discriminator is used to simultaneously improve the action value function and the imitator policy.
    \citet{56_opifvi_icra_2023} proposes learning an imitation policy from expert video demonstrations. It employs data augmentation to improve feature learning and train a GAIfO-style imitation policy. However, this approach maps high-dimensional pixel observations to a scalar adversarial reward, which overlooks the rich, dense information present in the image-based demonstrations. In contrast, \citet{57_patchail_iclr_2023} introduces a method that learns a multi-dimensional patch reward matrix, capturing the similarity between image patches in expert demonstrations and the imitator agent’s trajectory. This patch-level reward provides a more fine-grained and informative signal, which is then used to train a GAIfO-style adversarial imitation policy. Meanwhile, \citet{58_laifo_tmlr_2024} takes a different approach by first learning a latent space model. The expert video demonstrations are mapped into this latent space, and a GAIfO-style policy is trained using the latent representations rather than raw pixel observations. Most adversarial imitation learning approaches rely on standard neural network architectures, such as multilayer perceptrons (MLPs) or convolutional neural networks (CNNs), to train the discriminator. However, these discriminators often suffer from instability during training and require careful hyperparameter tuning. To address these challenges, \citet{55_difo_neurips_2024} proposes to use a diffusion model as the discriminator. Their results demonstrate that this approach provides a more stable and informative adversarial reward signal, leading to improved performance in imitator policy.
    One of the main limitations of the prior works is that it assumes that all expert state transitions are realizable, limiting the utilization of diverse expert demonstrations from different dynamical agents. \citet{33_ood_il_corl_2023, 50_fmdp_2021} addresses this challenge by identifying realizable expert state transitions using a discriminator network and then learning a GAIfO-style imitator policy. However, the work still assumes that the expert demonstrations are available from the expert's viewpoint, i.e., both the imitator and expert share the same state space. A more practical approach would involve collecting demonstrations from a third-person viewpoint. Studies by \citet{1_third_person_il_iclr_2017, 20_disentanGAIL_iclr_2021, 30_udil_neurips_2022} learn domain-invariant feature mapping by constraining mutual information and then subsequently learns a GAIfO-style policy over this feature space.

    \item \textit{Matching state occupancy measure}: The imitator policy of this class learns by minimizing the divergence between the state occupancy measures. A straightforward way to expand on this idea is to train a GAIfO-style policy with a discriminator trained to classify state visitation. To improve stability during adversarial training, the approach by \citet{12_sail_iclr_2020} replaces Jensen–Shannon loss with Wasserstein loss function \citep{wgan} to align with the expert behavior globally. It simultaneously trains an inverse dynamics model and uses BCO-style \citep{4_bco_ijcai_2018} training to align with the expert transitions locally. \citet{17_mobile_neurips_2021} proposes to learn a forward dynamics model and use it during adversarial training of the imitator policy. The uncertainty in predicting the next state using a forward dynamic model is used to incentivize exploration in those states. An alternative way, proposed by \citet{14_i2l_iclr_2020, 34_sail_ijcnn_2023}, uses discriminator to identify imitator transitions that exhibit similar expert behavior and then employs BCO \citep{4_bco_ijcai_2018} or AIRL \citep{airl_iclr_2018} style state-action based adversarial algorithm, respectively, to train imitator policy.
\end{enumerate}

\subsubsection{DICE based Imitation Learning}
DICE stands for stationary DIstribution Correction Estimation (DICE). In contrast to the methods discussed earlier, which directly train an imitator policy, the approaches in this section focus on optimizing the divergence between the stationary distribution within the distribution space. Subsequently, the estimated distribution is used to train the imitator policy.

\begin{enumerate}[label=\roman*]
    \item \textit{Matching state transition occupancy measure}: LobsDICE proposed by \citet{28_lobsdice_neurips_2022} focuses on the offline setting for LfO. It assumes that the imitator agent does not interact with the environment; instead, it has access to the action-labeled imitator transition data of arbitrary qualities along with the expert state-only demonstrations. The proposed algorithm optimizes inverse KL divergence between the state-transition distributions with an additional KL regularizer, which constrains the state-action distribution of the imitator policy, preventing it from deviating from the data support. This objective can be reduced to a single convex minimization problem that can be solved efficiently. The imitator policy is then obtained from the estimated imitator state-action distribution ($\mu^{\pi^\star}$) using behavior cloning.
    \[
        \min_\pi \mathop{\mathbb{E}}_{(s,a) \sim \mu^{\pi^\star}} [-\log \pi(a | s)]
    \]
    
    \item \textit{Matching state occupancy measure}: Similar to the work of LobsDICE \citep{28_lobsdice_neurips_2022}, SMODICE by \citet{24_smodice_icml_2022} and DILO by \citet{59_dilo_corl_2024} also focus on the offline setting for LfO and assume access to action-label imitator transition data. It proposes to optimize the upper bound of the inverse KL divergence between the state occupancy measures. However, unlike \citet{15_opolo_neurips_2020}, which optimizes the upper bound by solving minimax optimization, SMODICE reduces the objective to a single minimization problem. The estimated distribution is then used to train an imitator policy.
    \citet{60_pwdice_icml_2024} proposes a generalization of the SMODICE \cite{24_smodice_icml_2022} objective by unifying the $f$-divergence and Wasserstein distance metric. This unified framework enables a more accurate estimation of the imitator's state occupancy measure, resulting in improved imitation policy learning. 
    
\end{enumerate}

\section{Domain Adaptation}
\label{sec:domain_adaptation}

In this section, we explore the modifications made to LfO methods to incorporate demonstrations from different viewpoints or from non-identical agents. The central question we seek to answer is:

\begin{center}
    \textit{How are LfO algorithms modified to incorporate learning from demonstrations from different viewpoints or from non-identical agents?}
\end{center}

\subsection{Third Person Imitation Learning}
\label{sec:third_person_il}
Most of the works discussed in Section \ref{sec:algo_learn} assume that the demonstrations are in the first-person viewpoint. In this section, we will discuss approaches that learn to imitate from expert demonstrations when these demonstrations are available from different viewpoints and may be from morphologically different agents. The central idea is to bring the expert state or behavior information to the imitator's domain and then use the first-person imitation approach to learn to imitate the expert behavior. We broadly classify the methods based on how the knowledge is transferred from the source (expert) domain to the target (imitator) domain, as domain mapping and feature learning. 

\subsubsection{Domain Mapping} 
\begin{figure}
    \centering
    \includegraphics[width=0.9\textwidth]{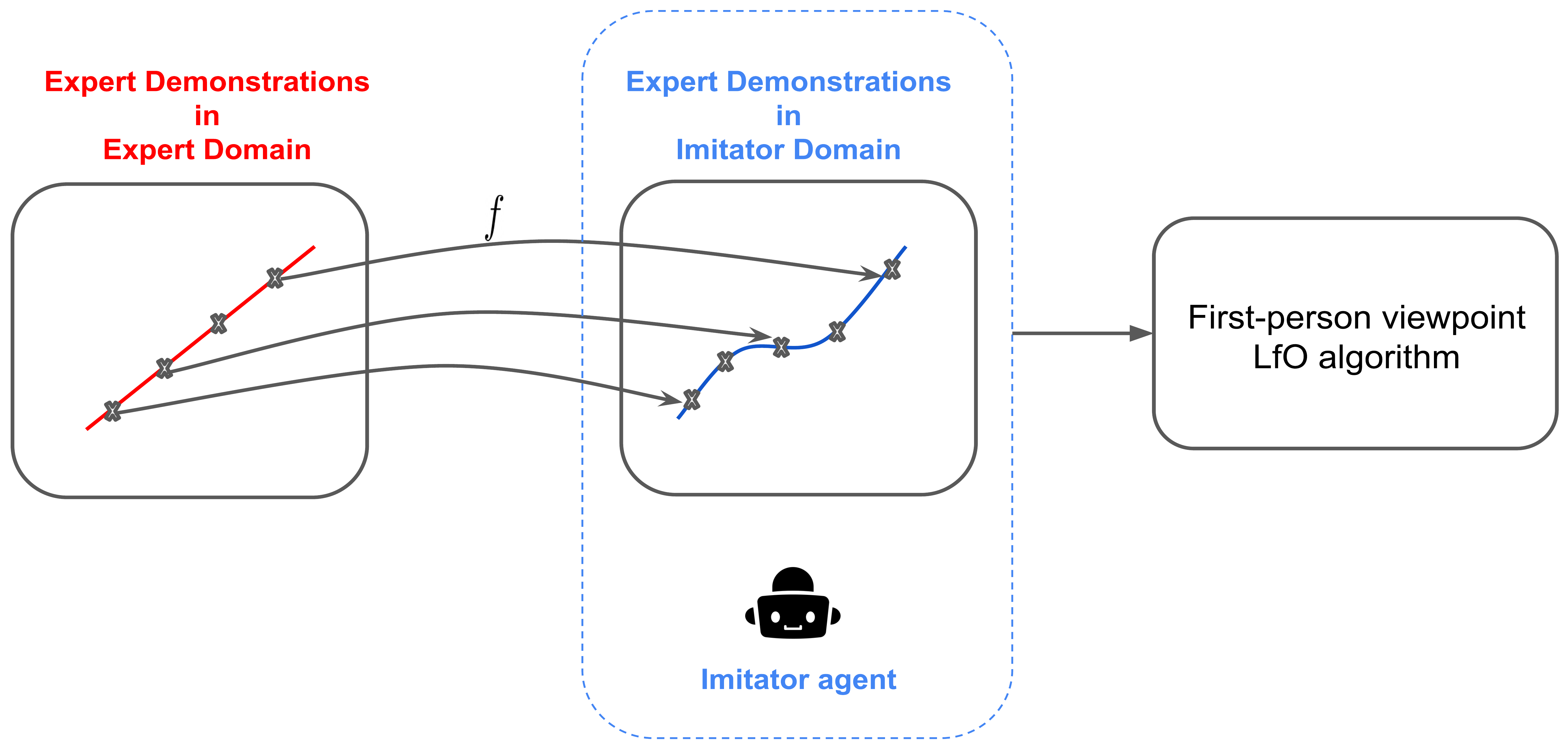}
    \Description{The diagram shows a mapping function \( f \) that translates expert demonstrations from expert domain into the imitator domain. These translated demonstrations are then used to train the imitator policy using a first-person LfO algorithm.}
    \caption{Domain adaptation using domain mapping. One can bridge the domain difference by learning a mapping function, $f$, to translate the expert demonstration to the imitator domain. Then, the translated expert demonstrations can be used to learn the imitator policy using a first-person LfO algorithm.}
    \label{fig:domain_mapping}
\end{figure}

The main objective of the works in this section is to learn a mapping function that can translate the expert behavior information into the imitator's domain. (See Figure \ref{fig:domain_mapping}) A simple approach would be to collect additional data in aligned pairs, where each pair consists of an expert state and its corresponding imitator state. By training a supervised mapping function, we can map the expert behavior in the imitator domain and use a first-person viewpoint approach to learn to imitate. However, obtaining such a state-aligned dataset (trajectories of random tasks collected in both the imitator and expert domain) is often impractical in most real-world scenarios. To address this issue, \citet{18_cross_domain_il_icml_2021} proposes to use unpaired and unaligned proxy trajectories to learn the mapping function. They utilize cycle consistency constraints on the expert and the imitator state space. In addition, they also enforce consistency in the temporal position of the states to align the trajectories across the two domains. Once the mapping function is learned, it can transfer the demonstrations from expert to imitator, and the imitator policy is learned via BCO-style \citep{4_bco_ijcai_2018} algorithm. The approach is limited to scenarios where one can easily collect proxy trajectories to learn the mapping function, and the mapping function may also end up encoding irrelevant information for the tasks. 
\citet{30_udil_neurips_2022} propose a method to learn task-relevant features in the expert domain. It achieves this by leveraging a mutual information criterion. The process involves collecting both expert state transitions and pseudo-random state transitions from the demonstrations. The model then learns a feature map by maximizing the mutual information between these state transitions and their labels, indicating whether the transitions are from expert or pseudo-random transitions. The imitator policy and the mapping from imitator to expert feature are jointly learned using an adversarial approach. An alternative approach proposed by \citet{2_imitation_from_raw_video_icra_2018} assumes that the latent context vector can translate expert domain information to the imitator domain. It learns a context translation model to covert expert state demonstrations to predict demonstrations in the imitator domain. The reward function is then defined as minimizing the Euclidean distance between the current imitator observation and the translated demonstration. This can be used to recover the imitator policy using the standard RL technique.  \citet{10_decoupled_hierarchical_il_neurips_2019} decouples the imitator policy into state planner and a control module. The state planner learns to predict the goal state in the imitator agent's viewpoint by learning a mapping function from expert domain to the imitator domain. The control module then takes the predicted goal state and the current state of the imitator agent to output sequence of actions that guide the agent towards the goal state.

\subsubsection{Feature Learning}
\begin{figure}
    \centering
    \includegraphics[width=0.9\textwidth]{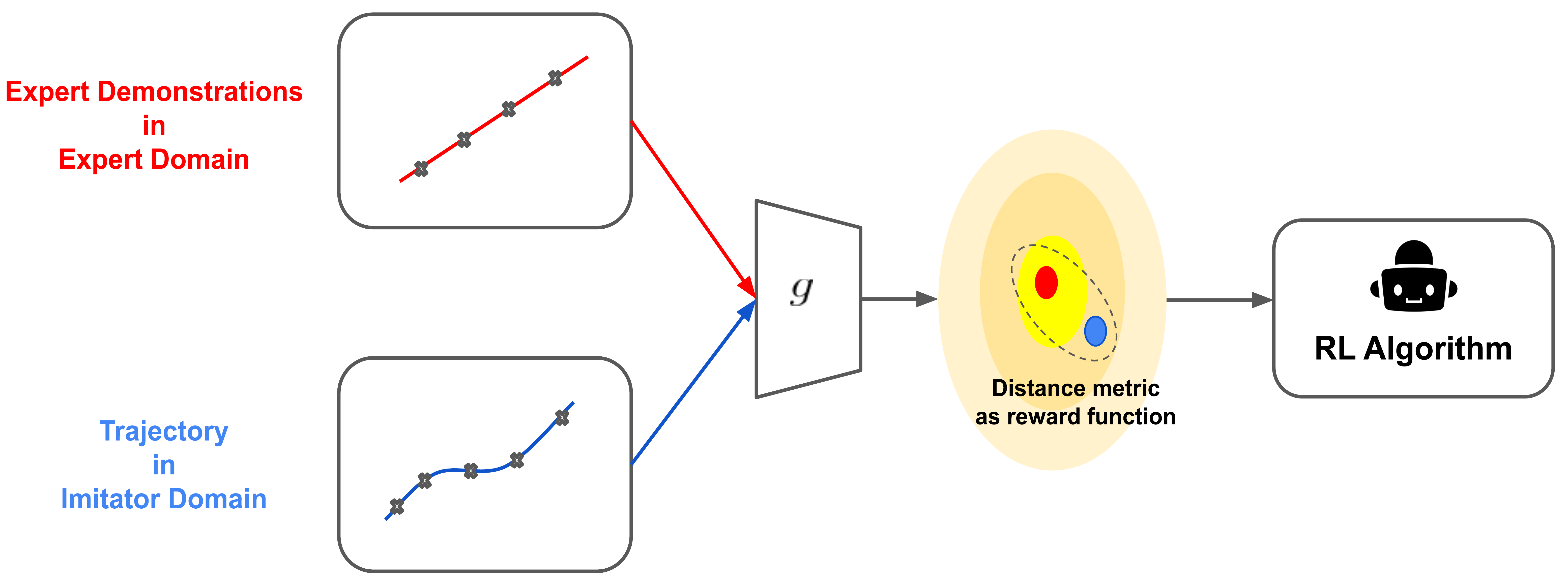}
    \Description{The diagram shows that the latent feature model $g$ maps the expert demonstrations and the imitator trajectories to a common latent space. A standard RL algorithm is used to learn an imitator policy that aims to bring the latent features of the expert demonstrations and the imitator trajectory close to each other.}
    \caption{Domain adaptation using feature learning. Works in this category learns a domain invariant latent feature model, $g$, that maps expert demonstrations and imitator trajectories to the same latent space. Different approaches impose different constraints on the latent space to learn the latent feature model. An imitator policy is then learned by a standard RL algorithm whose objective is to bring the latent features of the expert demonstrations and the imitator trajectory close.}
    \label{fig:feature_learning}
\end{figure}

Works in this category learn a domain invariant latent feature model that captures the behavior information of the expert and the imitator agent, as summarized in Figure \ref{fig:feature_learning}. \citet{42_tcn_icra_2018} learns viewpoint invariant feature representation from multiple temporally aligned expert demonstrations from different viewpoints. It achieves this by bringing multiple viewpoints of the same observation close in the embedding space while also ensuring that visually similar but functionally distinct temporal neighbors are separated. The reward function is then defined as the distance between the expert's feature and the observed imitator state feature representation. This reward function is used to recover the imitator policy using a standard RL algorithm. Collecting multiple temporally aligned expert demonstrations from different viewpoints can be a challenging task for complex environments. \citet{45_tdc_cmc_neurips_2018} addresses this limitation by introducing self-supervised objectives for learning latent features. Specifically, they propose two novel objectives: Temporal Distance Classification (TDC) and Cross-Modal Temporal Distance Classification (CMC). The TDC objective predicts the temporal distance between two observations within a single demonstration, while the CMC objective predicts the temporal distance between a given observation and the other modalities. The feature representation model learns meaningful abstractions of the environment dynamics by jointly optimizing these objectives. Similar to the previous approach, the reward function is defined as the distance between the expert and the imitator feature representations, and a standard RL algorithm is used to learn the imitator policy. Alternative approaches by \citet{1_third_person_il_iclr_2017} and  \citet{20_disentanGAIL_iclr_2021} propose to learn a feature representation model by constraining the mutual information (MI) between the observation representations and their corresponding domain labels, i.e whether these observations have arisen from an expert or the agent. Both works subdivide a discriminator into two distinct models: a feature extractor model and a domain-invariant discriminator model. However, the former work relies on low dimensional tasks and minimal discrepancy between the expert and imitator domains for a good performance. \citet{20_disentanGAIL_iclr_2021} alleviates the above problem, by proposing to add additional mutual information constraints. A \textit{goal completion} criterion is used to evaluate the progress to the target as demonstrated in the expert domain, which is used to tighten the MI constraint. Another constraint is added by considering the MI between the domain labels and randomly collected trajectories in the expert and agent domain to be close to zero. \citet{39_domain_adaptive_IL_neurips_2023} shows that disregarding domain information leads to loss of relevant information, and they include a cycle-consistency loss to include both the behavior as well as domain level information.

\subsection{Imitating non-identical agents}
\label{sec:imitating_non_identical_agents}
In this section we consider the case where there is a difference in the transition dynamics of the expert and the imitator agent. This arises in a wide range of real world problems; for example consider a human expert and an imitator robot. If the robot's actuator dynamics (e.g torque limits or friction) differs from that of the human's muscle dynamics, then there may be no action in some common state that can lead to the same next state transition. \citet{14_i2l_iclr_2020} proposes to solve the problem of the mismatch of dynamics by deriving two subproblems from upper bounding the max-entropy IRL objective. The first subproblem aims to minimize the divergence between the agent state distribution and some proxy state distribution, and the second aims to bring this proxy state distribution and the expert distribution closer. The two problems are solved using an adversarial style and using a Wasserstein critic respectively. This work aims to maximally follow the expert distribution; however since the dynamics are different not all expert demonstrations can be imitated by the imitator agent. \citet{51_imperfect_demo_icra_2021} learns an accurate inverse dynamics model that gives the correct action for a feasible state transition and outputs ``None'' for an infeasible state transition. They use this inverse dynamics model to identify non-transferable demonstrations and use only the transferable demonstrations to learn an imitator policy. However, learning an accurate inverse dynamics model can be difficult. \citet{50_fmdp_2021} proposes directly learning a feasibility score for the imitator agent based on the collected demonstrations. It constructs a feasibility MDP (f-MDP) based on the imitator MDP and the expert demonstrations. The optimal policy for the f-MDP maximally follows the expert behavior but is limited in the imitator environment. This optimal policy helps assign a feasibility score over the expert demonstrations. Finally, the imitator policy is learned by re-weighting the demonstrations by feasibility score. \citet{31_ailo_iclr_2022} proposes an algorithm that trains an intermediary policy in the imitator environment, which is then used as a surrogate expert to guide the training of the imitator agent. For the intermediary policy to be effective, the state transitions it generates in the imitator environment should be as close as possible to the state transitions in the expert demonstrations. To achieve this, the algorithm first learns a support function over expert state transitions using Random Estimation Distillation (RED) \cite{red}. This support function is then used to train the intermediary policy. \citet{qiu2023out} address the challenge of learning from a diverse set of expert demonstrations collected from multiple agents with different dynamics, which leads to a multimodal distribution of expert behaviors. This multimodality makes it difficult to learn a single unimodal policy. Unlike previous works, which overlook this issue, the authors introduce a sequence-based contrastive clustering approach to reduce the impact of multimodality and identifies relevant samples for learning a unimodal policy.

\section{LfO Resources}
\label{sec:lfo_resources}

\subsection{Datasets}
In this section, we discuss some common simulated and real datasets used in LfO. 
\subsubsection{Real World Datasets}
These datasets are collected in the real world by a human or a robot. \citet{mimedataset} dataset contains videos of various simple and complex activities demonstrated by a human as well as kinesthetically guided robotic demonstrations. The Multiview Pouring Dataset \cite{42_tcn_icra_2018} contains demonstrations of pouring liquid in a cup in a first-person and third-person viewpoint. \citet{activitynet} dataset contains a wide array of day-to-day human activity demonstrations in the form of videos with a focus on complex tasks.
\subsubsection{Simulated Datasets}
D4RL \cite{d4rl} and Robomimic \cite{robomimic} are widely used benchmarks for offline reinforcement learning, containing trajectories of varying optimality. Most works in this survey train an expert using a standard RL algorithm, which is used to collect expert trajectories across a diverse set of environments. These include those with discrete (Classic Control) and continuous (Mujoco \cite{mujoco}, Brax \cite{brax2021github}) action spaces available in the OpenAI Gym \cite{gym} or Gymnasium \cite{towers2024gymnasium} frameworks. Additionally, some papers also performed experiments on Mujoco-based DM Control Suite \cite{dmcontrol} and other miscellaneous benchmarks like VizDoom \cite{Wydmuch2019ViZdoom} a discrete action space environment based on the Doom game, XMagical benchmark \cite{toyer2020magical} a continuous environment to investigate third-person visual imitation learning and Atari \cite{atari} a discrete environment simulated via the Arcade Learning Environment \cite{bellemare13arcade}.

\subsection{Implementations}
We provide a comprehensive list of implementations of some important works covered in this article in Table \ref{tab:lfo-resources}. We adhere to our classification discussed in Section \ref{sec:algo_learn} along with specifying whether these algorithms can be used to address the different domain shift scenarios (third person imitation learning and learning from different agents) discussed in Section \ref{sec:domain_adaptation}, that are more common and practical in the real world. We also provide information on whether the official implementation is available or not. 

\begin{table}[!tp]
\begin{center}
  \addtolength{\tabcolsep}{-3pt}
  \def\arraystretch{1.5}
  \fontsize{9.0pt}{10.25pt}\selectfont
  \caption{We provide a list of algorithms with their ability to handle domain adaptation and their implementations. Implementations marked \text{\cmark} and \text{\xmark} depict whether there exists an official implementation of the algorithm or not. Third party implementations could be found for the algorithms, but they are upto the users' discretion.}
  \label{tab:lfo-resources}
\begin{tabular}{c l cc c}
\hline
\rowcolor[HTML]{EFEFEF} 
\cellcolor[HTML]{EFEFEF}                                          & \multicolumn{1}{c|}{\cellcolor[HTML]{EFEFEF}}                                     & \multicolumn{2}{c|}{\cellcolor[HTML]{EFEFEF}\textbf{Domain Adaptation}}                                                                                                  & \cellcolor[HTML]{EFEFEF}                                          \\ \cline{3-4}
\rowcolor[HTML]{EFEFEF} 
\multirow{-2}{*}{\cellcolor[HTML]{EFEFEF}\textbf{Classification}} & \multicolumn{1}{c|}{\multirow{-2}{*}{\cellcolor[HTML]{EFEFEF}\textbf{Algorithm}}} & \multicolumn{1}{c|}{\cellcolor[HTML]{EFEFEF}Section \ref{sec:third_person_il}} & \multicolumn{1}{c|}{\cellcolor[HTML]{EFEFEF}Section \ref{sec:imitating_non_identical_agents}} & \multirow{-2}{*}{\cellcolor[HTML]{EFEFEF}\textbf{Implementation}} \\ \hline
                                                                  & \citet{4_bco_ijcai_2018}                                           & \multicolumn{1}{c|}{\xmark}                                                        & \xmark                                                & \xmark                                             \\ \cline{2-5} 
                                                                  & \citet{29_ild_cvpr_2022}                                           & \multicolumn{1}{c|}{\xmark}                                                        & \xmark                                                & \cmark                                             \\ \cline{2-5} 
\multirow{-3}{*}{Supervised}                                      & \citet{11_ilpo_icml_2019}                                         & \multicolumn{1}{c|}{\cmark}                                                        & \xmark                                                & \cmark                                             \\ \hline
                                                                  & \citet{3_gsp_iclr_2018}                                            & \multicolumn{1}{c|}{\xmark}                                                        & \xmark                                                & \cmark                                             \\ \cline{2-5} 
                                                                  & \citet{7_silo_corl_2019}                                           & \multicolumn{1}{c|}{\xmark}                                                        & \xmark                                                & \xmark                                             \\ \cline{2-5} 
                                                                  & \citet{10_decoupled_hierarchical_il_neurips_2019}                & \multicolumn{1}{c|}{\cmark}                                                        & \xmark                                                & \cmark                                             \\ \cline{2-5} 
\multirow{-4}{*}{Goal From Expert Demonstration}                  & \citet{27_depo_icml_2022}                                          & \multicolumn{1}{c|}{\xmark}                                                        & \xmark                                                & \cmark                                             \\ \hline
                                                                  & \citet{2_imitation_from_raw_video_icra_2018}                    & \multicolumn{1}{c|}{\cmark}                                                        & \xmark                                                & \cmark                                             \\ \cline{2-5} 
                                                                  & \citet{45_tdc_cmc_neurips_2018}                                   & \multicolumn{1}{c|}{\cmark}                                                        & \xmark                                                & \xmark                                             \\ \cline{2-5} 
                                                                  & \citet{42_tcn_icra_2018}                                           & \multicolumn{1}{c|}{\cmark}                                                        & \xmark                                                & \xmark                                             \\ \cline{2-5} 
                                                                  & \citet{26_bootifol_neurips_workshop_2022}                         & \multicolumn{1}{c|}{\xmark}                                                        & \xmark                                                & \cmark                                             \\ \cline{2-5} 
                                                                  & \citet{44_osl_icra_2019}                                           & \multicolumn{1}{c|}{\xmark}                                                        & \xmark                                                & \cmark                                             \\ \cline{2-5} 
                                                                  & \citet{19_form_icml_2021}                                          & \multicolumn{1}{c|}{\xmark}                                                        & \xmark                                                & \xmark                                             \\ \cline{2-5} 
\multirow{-7}{*}{Reward Engineering}                              & \citet{5_trex_icml_2019}                                           & \multicolumn{1}{c|}{\xmark}                                                        & \xmark                                                & \cmark                                             \\ \hline
                                                                  & \citet{9_gaifo_icml_workshop_2019}                                & \multicolumn{1}{c|}{\xmark}                                                        & \xmark                                                & \xmark                                             \\ \cline{2-5} 
                                                                  & \citet{6_iddm_neurips_2019}                                        & \multicolumn{1}{c|}{\xmark}                                                        & \xmark                                                & \xmark                                             \\ \cline{2-5} 
                                                                  & \citet{25_robust_gaifo_aamas_2022}                                & \multicolumn{1}{c|}{\cmark}                                                        & \xmark                                                & \cmark                                             \\ \cline{2-5} 
                                                                  & \citet{15_opolo_neurips_2020}                                      & \multicolumn{1}{c|}{\xmark}                                                        & \xmark                                                & \cmark                                             \\ \cline{2-5} 
                                                                  & \citet{33_ood_il_corl_2023}                                       & \multicolumn{1}{c|}{\xmark}                                                        & \cmark                                                & \cmark                                             \\ \cline{2-5} 
                                                                  & \citet{50_fmdp_2021}                                                & \multicolumn{1}{c|}{\xmark}                                                        & \cmark                                                & \xmark                                             \\ \cline{2-5} 
                                                                  & \citet{1_third_person_il_iclr_2017}                              & \multicolumn{1}{c|}{\cmark}                                                        & \xmark                                                & \cmark                                             \\ \cline{2-5} 
                                                                  & \citet{20_disentanGAIL_iclr_2021}                                  & \multicolumn{1}{c|}{\cmark}                                                        & \xmark                                                & \cmark                                             \\ \cline{2-5} 
                                                                  & \citet{30_udil_neurips_2022}                                       & \multicolumn{1}{c|}{\cmark}                                                        & \xmark                                                & \xmark                                             \\ \cline{2-5} 
                                                                  & \citet{12_sail_iclr_2020}                                          & \multicolumn{1}{c|}{\cmark}                                                        & \xmark                                                & \cmark                                             \\ \cline{2-5} 
                                                                  & \citet{17_mobile_neurips_2021}                                     & \multicolumn{1}{c|}{\xmark}                                                        & \xmark                                                & \cmark                                             \\ \cline{2-5} 
                                                                  & \citet{14_i2l_iclr_2020}                                           & \multicolumn{1}{c|}{\xmark}                                                        & \cmark                                                & \cmark                                             \\ \cline{2-5} 
\multirow{-13}{*}{Distribution Matching}                          & \citet{34_sail_ijcnn_2023}                                         & \multicolumn{1}{c|}{\xmark}                                                        & \xmark                                                & \cmark                                             \\ \hline
                                                                  & \citet{28_lobsdice_neurips_2022}                                   & \multicolumn{1}{c|}{\xmark}                                                        & \xmark                                                & \xmark                                             \\ \cline{2-5} 
\multirow{-2}{*}{DICE Algorithms}                                 & \citet{24_smodice_icml_2022}                                       & \multicolumn{1}{c|}{\xmark}                                                        & \xmark                                                & \cmark                                             \\ \hline
\end{tabular}
\end{center}
\end{table}

\section{Connection with other related fields}
\label{sec:other_fields}

This section will discuss how ideas from other fields are connected or can help improve the design of LfO algorithms. 

\subsection{Offline RL}
In many settings, online interactions can be expensive (e.g., robotics) or dangerous (e.g., autonomous driving or surgical robotics). Moreover, even in domains where online interactions are feasible, we may still prefer to use existing data to achieve effective generalization. However, in LfO, online interactions are required to address the challenge of the distribution shift \citep{bc_covariate}. When the policy transitions into states outside the training dataset (expert demonstrations), the learned policy may take sub-optimal actions and remain in the out-of-distribution states for the remainder of the trial. As a result, our learned policy may deviate from the expert behavior. Another important reason for online interactions is that our expert demonstrations contain state-only information. Hence, the imitator policy also needs to learn the effect of the action on the imitator agent in order to mimic the expert's behavior. Therefore, offline IL requires an additional proxy state-action transition dataset from the imitator agent and expert demonstrations.  \citet{24_smodice_icml_2022, 28_lobsdice_neurips_2022} methods utilize bellman flow constraints (equation: \ref{eq:bellman_flow}) and exploit the convex duality to obtain density-ratio between the imitator and the proxy data occupancy measures. It then uses a behavior cloning style algorithm to obtain the imitator policy.

We note that this is a developing area and ideas from offine RL/IL \citep{offline_rl_sergey, milo_neurips_2021, safedice_neurips_2023, mahalo_icml_2023} can also help improve this field of works.

\subsection{Model-Based RL}

Model-based RL \citep{model_based_rl_survey_2023} refers to a broad class of methods that utilizes either a known or learned dynamics model. Some commonly used model-based RL algorithm learns the \textit{forward dynamics model} and then use it for lookahead planning, often using model-predictive control (MPC) \citep{tassa2012synthesis, guided_policy_search_neurips_2014} or MCTS \citep{efficient_zero_neurips_2021}. Others utilize the forward dynamics model as a proxy for the environment to generate synthetic samples to augment the samples available for model-free RL methods \citep{dyna_q_acm_1991}. \citet{RRT_report_1998} uses an \textit{inverse dynamics model} for path planning. Model-based RL is also susceptible to distribution shifts. When a policy is learned by exploiting the dynamics model, the policy may produce out-of-distribution states and actions at which the model may erroneously predict the next states rather than the actual state obtained from the real environment. This model exploitation can lead to poor performance in the real environment \citep{model_based_rl_survey_2023}.  

In LfO, \citet{17_mobile_neurips_2021} uses the forward dynamics model as a proxy for the environment to obtain synthetic samples to train the imitator policy. It additionally uses uncertainty estimates during policy learning to address the issue of distribution shift. \citet{3_gsp_iclr_2018, 11_ilpo_icml_2019} uses the forward dynamics model, and \citet{12_sail_iclr_2020, 15_opolo_neurips_2020} uses the inverse dynamics model as a regularizer to constraint imitator policy learning. \citet{4_bco_ijcai_2018} proposes to use the inverse dynamics model. It uses the inverse model to estimate the actions for the given expert state transitions. Subsequently, a behavior-cloned policy is trained based on the expert state and the predicted action. The proposed algorithm is susceptible to distribution shift. The inverse dynamics model may predict incorrect actions for some expert state transitions. Training the imitator policy with these erroneous action values may lead to poor performance in the real environment. \citet{10_decoupled_hierarchical_il_neurips_2019, 27_depo_icml_2022} uses the inverse dynamics model as part of the imitator policy. It subdivides the imitator policy into a state-prediction model, which predicts the next goal state to visit, and a control module, an inverse dynamics model that predicts sequence of action given the current state and goal state.


\subsection{Hierarchical RL}
Hierarchical Reinforcement Learning (HRL) \citep{hrl_survey} decomposes the long-horizon RL task into a hierarchy of subtasks. The higher-level policy selects subtasks as its actions, while the lower-level policy learns to perform each subtask using internal rewards  \citep{7_silo_corl_2019} or by minimizing an objective function \citep{3_gsp_iclr_2018, 10_decoupled_hierarchical_il_neurips_2019} related with that subtask. In goal-conditioned HRL, the subtask’s objective is to choose a goal state for the low-level policy to reach. The subtask receives positive feedback if the selected goal state contributes to task completion. Meanwhile, the low-level policy receives feedback when the agent reaches the specified goal state.

\citet{3_gsp_iclr_2018} uses predefined subgoals that select consecutive expert states from the demonstration as the goal for the low-level policy to reach. The low-level policy then utilizes the current and goal states to predict a sequence of primitive actions. This low-level policy can be trained offline in a supervised manner using any transitions. \citet{10_decoupled_hierarchical_il_neurips_2019} also uses a similar low-level policy; however, the high-level policy learns to predict the goal state given the current state and the expert demonstrations. \citet{7_silo_corl_2019}, on the other hand, formulates both learning the subgoals and the low-level policy as a goal-conditioned HRL problem.

\section{Open Problems and Future Directions}
\label{sec:open_problems}

Imitation learning draws inspiration from how humans learn to perform tasks by observing others. In this section, we highlight several research directions that have the potential to significantly advance the field.

\begin{itemize}
    \item \textbf{Learning from experts with varying skills}: This has been explored in several works \citep{20_disentanGAIL_iclr_2021, 50_fmdp_2021, 51_imperfect_demo_icra_2021, 31_ailo_iclr_2022, 30_udil_neurips_2022, 33_ood_il_corl_2023}. However, these approaches often involve few experts with specific skills. The recent advancement in the foundation model \citep{ving_icra_2021, open_x_embodiment_rt_x_2023} for robotics can be pretrained with a large number of experts with diverse skills. These models have demonstrated remarkable success in handling downstream applications. However, these models require trajectories in the form of state and action pairs. If we can adapt them to learn from state-only trajectories, they can be trained with an even larger dataset, as collecting state-only trajectories are easier.

    \item \textbf{Third person imitation learning}: Many existing methods either trim the irrelevant parts from the third-person demonstrations \citep{42_tcn_icra_2018} or can only handle demonstrations with minor changes in camera angle \citep{1_third_person_il_iclr_2017}. However, if we want to utilize internet scale passive resources like YouTube videos, we may want to explore IL when the expert demonstrations are available in a third-person viewpoint.

    \item \textbf{Subgoal identification from state-only demonstrations}: Decoupling imitator policy into goal state prediction modules and a low-level controller module can help train both modules independently. Low-level controller can be learned independent of the task. Most existing work in the literature focuses on learning from experts, similar to the imitator agent. Extending this work to identify better subgoal states can help improve the ability to learn from morphologically different expert demonstrations.

    \item \textbf{Plan to imitate}: Model-based IL holds promise for enhancing sample efficiency. Existing approaches often treat the model as a proxy for the environment to collect synthetic transitions \citep{17_mobile_neurips_2021} or as a regularizer \citep{12_sail_iclr_2020, 15_opolo_neurips_2020} for the imitator policy. However, recent successes in model-based RL have emerged due to using the model as a planner \citep{when_to_use_model_neurips_2019, muzero_nature_2020, efficient_zero_neurips_2021}. It draws inspiration from how humans use mental models to plan their actions. Leveraging the model as a planner in imitation learning could help improve sample efficiency.

    \item \textbf{Safe Imitation}: Most research in imitation learning centers on replicating expert behavior. However, in real-world scenarios, adhering to safety constraints often takes precedence over merely copying the expert's behavior. It is common to see humans imitating others while prioritizing their safety. Advancing research in safe imitation learning could lead to greater success in practical applications.

    \item \textbf{Evaluation Metric}: Most studies evaluate the performance of the imitator agent by measuring the return at the end of an episode, using this as an indicator of how well the imitator has replicated the expert's behavior. However, this performance measure does not reflect imitation, as the imitator may achieve a similar return through different behaviors. Developing a better evaluation metric is essential for effectively assessing the imitation behavior produced by novel methods and is crucial for maintaining progress in the field.
\end{itemize}

\section{Related Surveys}
\label{sec:related_works}
Imitation learning has been successful and has received significant attention in recent years \citep{IL_survey_hussein_2017, algorithmic_IL_survey_2018, current_IL_survey_2022, current_IL_survey_zare_2023, LfO_survey}. However, the survey in the field of LfO lacks literature. \citet{IL_survey_hussein_2017} provides a comprehensive review of the design choices available at each stage of imitation learning, while the \citet{algorithmic_IL_survey_2018} focuses on the algorithmic aspects of IL. Despite their thoroughness, both works primarily review literature where actions are part of the demonstrations, hence lacking coverage on LfO. \citet{current_IL_survey_2022, current_IL_survey_zare_2023} provides a survey of recent progress in IL. However, their works neither offer a complete overview nor incorporate the recent literature on LfO. 

\citet{LfO_survey} proposes a novel taxonomy for LfO. This taxonomy broadly classifies LfO algorithms into two categories: model-based and model-free approaches. In model-based LfO methods, the algorithms learn a dynamics model during the training of the imitation policy. This dynamics model can be a \textit{forward dynamics model} or an \textit{inverse dynamics model}. On the other hand, in model-free approaches, the imitation policy is either trained through \textit{reward engineering} or via \textit{adversarial training}. We found that the recent state-only imitation learning algorithms may not neatly fit into the established taxonomy. For instance, \citet{11_ilpo_icml_2019} employs both \textit{forward} and \textit{inverse} dynamics models to learn to imitate expert behavior, while \citet{27_depo_icml_2022} combines an \textit{inverse dynamics model} with an \textit{adversarial-style} approach. Additionally, it is challenging to categorize DICE-based approaches \citep{28_lobsdice_neurips_2022, 24_smodice_icml_2022} within the framework described by \citet{LfO_survey}. Therefore, there is a need for a dedicated survey that addresses the advancements and the unique challenges in the field of LfO.
\section{Summary}
\label{sec:summary}
This article introduces a novel taxonomy and employs it to survey recent works on LfO. We discussed works based on how trajectory datasets are constructed and how algorithms are designed to handle different trajectory datasets. We also discussed the assumptions, strengths, and limitations of these algorithms. Additionally, we highlighted how ideas from other related fields have been used to drive progress in LfO. Finally, we suggest some open problems in the field of LfO, which can provide valuable insights for future research directions.

\begin{acks}
The authors thank Prof. Matthew E. Taylor for providing helpful feedback on our article. Returaj Burnwal acknowledges financial support from the Prime Minister's Research Fellowship (PMRF), Ministry of Education, Government of India.
\end{acks}

\bibliographystyle{ACM-Reference-Format}
\bibliography{references}

\end{document}